\documentclass[journal]{IEEEtran}
\PassOptionsToPackage{numbers}{natbib}
\usepackage[cmex10]{amsmath}
\usepackage{amssymb}
\usepackage{algorithmic}
\ifCLASSOPTIONcompsoc
 \usepackage[caption=false,font=normalsize,labelfont=sf,textfont=sf]{subfig}
\else
 \usepackage[caption=false,font=footnotesize]{subfig}
\fi
\usepackage{url}
\usepackage{graphicx}
\usepackage[table,xcdraw]{xcolor}
\usepackage{caption}
\usepackage{colortbl}
\usepackage{booktabs}     
\usepackage{siunitx}
\usepackage{multirow}     
\usepackage{xcolor}
\usepackage{etoolbox}
\newrobustcmd\B{\DeclareFontSeriesDefault[rm]{bf}{b}\bfseries}
\DeclareUnicodeCharacter{2212}{-}
\usepackage[flushleft]{threeparttable}

\begin{document}
\title{
SSVEP-DAN: Data Alignment Network for SSVEP-based Brain Computer Interfaces
}

\author{
Sung-Yu Chen, 
Chi-Min Chang, 
Kuan-Jung Chiang, 
and Chun-Shu Wei,~\IEEEmembership{Member,~IEEE}
\thanks{This work was supported in part by the National Science and Technology Council (NSTC) under Contracts 109-2222-E-009-006-MY3, 110-2221-E-A49-130-MY2, 110-2314-B-037-061, and 112-2222-E-A49-008-MY2; and in part by the Higher Education Sprout Project of National Yang Ming Chiao Tung University and Ministry of Education. Corresponding author: Chun-Shu Wei (wei@nycu.edu.tw).}
\thanks{
Sung-Yu Chen, Chi-Min Chang, and Chun-Shu Wei are with the Department of Computer Science, National Yang Ming Chiao Tung University (NYCU), Hsinchu, Taiwan. Chun-Shu Wei is also with the Institute of Education and the Institute of Biomedical Engineering, NYCU, Hsinchu, Taiwan.}
\thanks{
Kuan-Jung Chiang is with the Arctop Inc., CA, USA}
}

\maketitle

\begin{abstract}

Steady-state visual-evoked potential (SSVEP)-based brain-computer interfaces (BCIs) offer a non-invasive means of communication through high-speed speller systems. However, their efficiency heavily relies on individual training data obtained during time-consuming calibration sessions. To address the challenge of data insufficiency in SSVEP-based BCIs, we present SSVEP-DAN, the first dedicated neural network model designed for aligning SSVEP data across different domains, which can encompass various sessions, subjects, or devices. Our experimental results across multiple cross-domain scenarios demonstrate SSVEP-DAN's capability to transform existing source SSVEP data into supplementary calibration data, significantly enhancing SSVEP decoding accuracy in scenarios with limited calibration data. We envision SSVEP-DAN as a catalyst for practical SSVEP-based BCI applications with minimal calibration. The source codes in this work are available at: https://github.com/CECNL/SSVEP-DAN.

\end{abstract}


%
\IEEEpeerreviewmaketitle

\section{Introduction}
\IEEEPARstart{S}teady-State Visual Evoked Potentials (SSVEPs) are a specific type of EEG signal that occurs in the cortical region of the human brain when an individual focuses their attention on visual stimuli flickering at specific frequencies \cite{david1977ssvep1, norcia2015ssvep2}. SSVEPs are known for their robustness  \cite{waytowich2016multiclass, waytowich2018compact} and have emerged as a reliable control signal for non-invasive brain-computer interfaces (BCIs) \cite{wolpaw2007brain, birb2006breaking}, facilitating communication between individuals and computers or external devices. The utilization of SSVEP-based BCIs has found widespread applications in various practical domains, including spelling \cite{cheng2002design, chen2015high}, gaming \cite{martivsius2016prototype, chen2017single}, and device control \cite{muller2011using, guneysu2013ssvep}.

\begin{figure}[t!]
    \centering
    \includegraphics[width=\columnwidth,trim={0cm 0cm 0cm 0cm},clip]
    {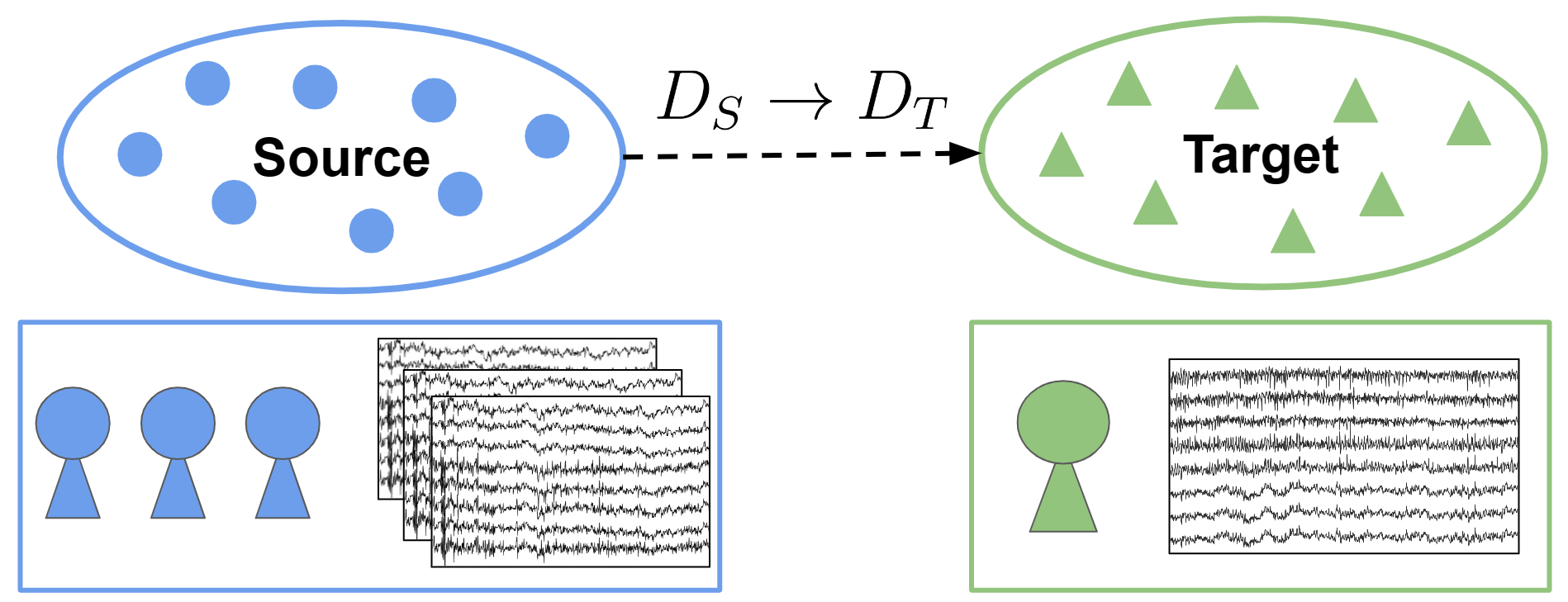}
    \caption{An illustration of domain adaptation for boosting the calibration of SSVEP-based BCI. Transferring existing data from the source subjects ($D_S$) to the target subject ($D_T$) provides additional calibration for a target user and therefore reduce the required amount of data from individual calibration.}
    \label{fig:DA}
\end{figure}

\begin{figure}[b!]
    \centering
    \includegraphics[width=\columnwidth]{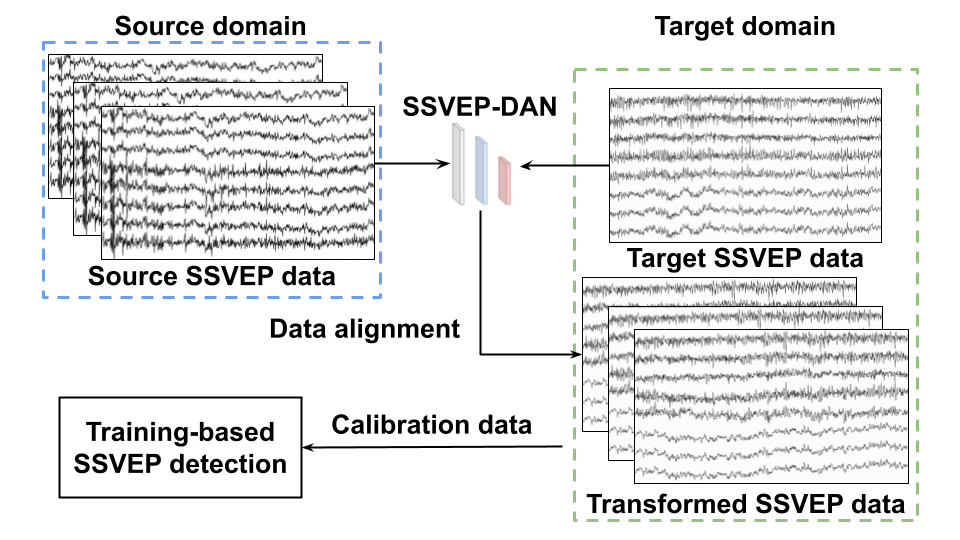}
    \caption{
    The proposed SSVEP data alignment framework based on SSVEP-DAN. The SSVEP-DAN learns the transformation between the source and target domain through minimizing the difference between the transformed and target SSVEP data. Then, the transformed SSVEP data, together with the target SSVEP data, serve as the calibration data for the training-based SSVEP detection.
    }
    \label{fig:DA_framework}
\end{figure}

To accurately detect and analyze users' SSVEPs for distinguishing corresponding stimuli, the development of efficient decoding algorithms has become significantly important.
Canonical correlation analysis (CCA) \cite{lin2007cca, bin2009onelineCCA} has been proposed as a training-free technique used to assess the relationships between multichannel SSVEP and the reference sinusoidal signals corresponding to each stimulation frequency. While the performance of training-free methods is limited by the individual differences, training-based SSVEP detection leverages individual calibration data to improve the performance. The most widely-used training-based SSVEP detection algorithm is task-related component analysis (TRCA) \cite{nakan2018trca, chiang2022fastTRCA}, aiming to separate task-related from non-task-related information by maximizing SSVEP data reproducibility within each trial. As the gold standard approach to online SSVEP-based BCI \cite{ladouce2022improving, jiang2022user, bai2023hybrid}, the success of TRCA has inspired the development of novel neural-network-based SSVEP detection models such as Conv-CA \cite{li2020conCCA} and bi-SiamCA \cite{zhang2022bidirectional}.
However, the calibration process for collecting individual data is often time-consuming and laborious, leading to significant visual fatigue in subjects \cite{cao2014fatigue}. Furthermore, due to substantial variability between subjects, simply concatenating training data from a larger pool of participants can potentially lead to a decrease in the decoding algorithm's performance \cite{Chiang2021LST}. This technical challenge manifests as a domain adaptation problem, a subcategory of transfer learning aiming to transfer knowledge from a source domain to improve a model's performance on a target domain \cite{pan2010survey}. In this study, we develop a domain adaptation technique to mitigate inter-domain disparities and enhance SSVEP decoding algorithm performance, particularly in scenarios with limited calibration data. Figure \ref{fig:DA} illustrates the proposed framework, where the target domain simulates a new user of the SSVEP-based BCI, and the source domain represents subjects providing existing SSVEP data.

\begin{table*}[t!]
  \caption{Related work on the domain adaptation methods in SSVEP studies.}
  \centering
\resizebox{\textwidth}{!}{%
\begin{tabular}{llcclccc}
\toprule
\multicolumn{5}{c}{} &
\multicolumn{3}{c}{Domain transferred} \\
\cmidrule(r){6-8}
Method & DA approaches & Share subspace & stimulus-specific & Classifier input & Subjects & Sessions & Devices \ \\
\midrule
RPA & Subspace alignment & Yes & No & SPD matrix & Yes  & Yes & No  \\
SLR  & Subspace alignment & Yes & Yes & Spatial patterns & Yes & Yes & Yes \\
ALPHA  & Subspace alignment & Yes & Yes & Spatial patterns & Yes & Yes & Yes\\
TSA  & Subspace alignment & Yes & No & Tangent vectors &Yes & Yes & Yes \\
LST  & Data alignment & No & Yes & Time-series data & Yes & Yes & Yes\\
\bottomrule
\end{tabular}}
  \label{tab:related_work}
\end{table*}

\begin{figure*}[!ht]
    \centering
    \includegraphics[width=\textwidth]{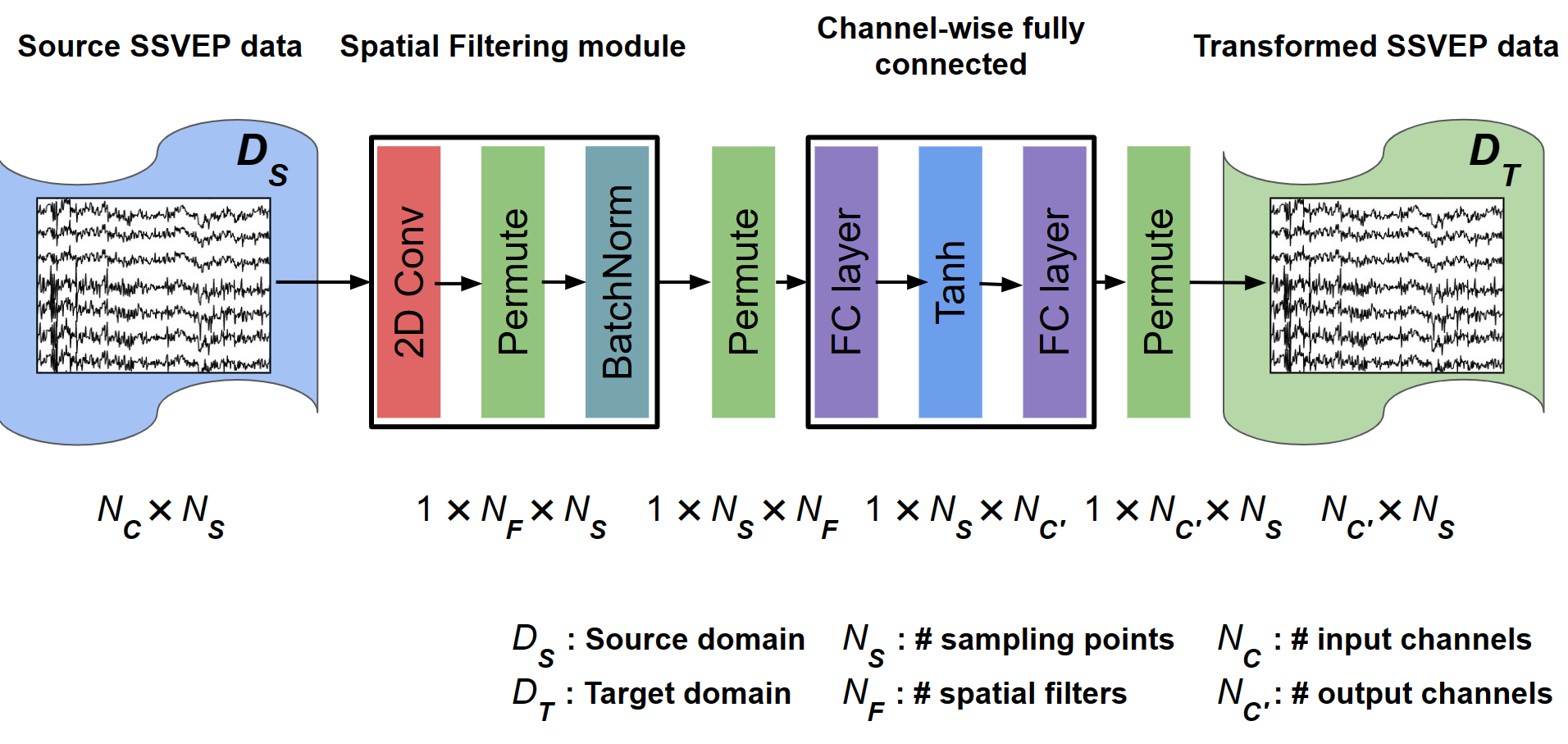}
    \captionsetup{justification=centering}
    \caption{The architecture of the proposed SSVEP-DAN.}
    \label{fig:Network}
\end{figure*}

Recently, numbers of studies have applied domain adaptation techniques to reduce the calibration effort or enhance the detection performance in SSVEP-based BCIs \cite{Chiang2021LST, rodrigues2018rpa, masaki2020facil, liu2022alpha, bleuze2022tsa}. A majority of domain adaptation methods focus on aligning data from different participants to a common subspace to reduce inter-subject data distribution differences. However, these methods have limitations in their compatibility with subsequent SSVEP detection algorithms due to their non-SSVEP signal outputs, such as covariance matrices \cite{rodrigues2018rpa}, tangent vectors \cite{bleuze2022tsa}, or spatial patterns \cite{liu2022alpha, masaki2020facil}. Therefore, each subspace alignment method requires a dedicated framework of SSVEP detection that poses challenge in combining with commonly used SSVEP detection algorithms based on conventional techniques (e.g., TRCA) or deep-learning approaches (e.g., Conv-CA). 
In contrast, data alignment performs adaptation between samples of the source domain and the target domain and generates time-series SSVEP signals as an output. A recently introduced data alignment technique, known as least-square transformation (LST) \cite{Chiang2021LST}, has the capacity to transmute the SSVEP waveform from the source subjects into supplementary calibration data for a target subject without any interference when integrated with subsequent SSVEP detection algorithms.
Nonetheless, the stimulus-dependent training process of LST based on individual visual stimulus restricts the quantity of available data during model fitting and could lead to redundant and invalid transformation models. Recent studies suggest that SSVEP signals exhibit non-linear characteristics \cite{aznan2019simulating, kwon2022novel, pan2023short}, yet LST's capabilities are confined to linear transformations, thereby exhibiting limited efficacy in accommodating non-linear transformations and noise tolerance.

To address the issues of current domain adaptation techniques for SSVEP detection, we introduce a neural network-based data alignment method named SSVEP-DAN. SSVEP-DAN features non-linear mapping of source SSVEP signals to generate desirable target domain data, with a fitting process allowing stimulus-independent training to acquire a relatively robust transformation.
The transformed SSVEP signals are readily used as additional calibration data for the target subject with any training-based SSVEP detection algorithm. 
This work introduces the first dedicated neural network architecture for SSVEP data alignment, as illustrated in Figure \ref{fig:DA_framework}. The proposed architecture integrates pioneering training methodologies, including stimulus-independent training and pre-training techniques, formulated to address the challenges posed by data scarcity. The viability and effectiveness of the proposed SSVEP-DAN framework are evaluated in conjunction with the most common high-speed SSVEP decoding algorithm, TRCA \cite{nakan2018trca, wong2020spatial, chiang2022fastTRCA}, through a series of experiments encompassing diverse cross-domain adaptations that correspond to scenarios of practical SSVEP-based BCI applications.

\section{Related Work}

Domain adaptation techniques aim to adapt the trained model from the source domain to the target domain by leveraging the available data from the target domain while utilizing the knowledge learned from the source domain. The goal is to improve the model's performance and generalization capabilities across different users without the need for extensive re-training or user-specific calibration.
This section provides the background of domain adaptation for SSVEP-based BCI and reviews relevant studies on existing domain adaptation approaches as summarized in Table \ref{tab:related_work}.

Domain adaptation methods in SSVEP-based BCIs are typically categorized into two groups: 1) subspace alignment, which involves alignment between domains of feature spaces or subspaces, and 2) data alignment, which performs alignment on the SSVEP signals between domains based on their transformation relationships \cite{sarafraz2022domain, wan2021review, wu2020transfer}.

\subsection{Subspace Alignment}
Subspace alignment methods align source and target domains of feature subspace to a common subspace where the the discrepancy between the two domains reduces. Riemannian Procrustes Analysis (RPA) \cite{rodrigues2018rpa} achieves this by applying simple geometric transformations (translation, scaling, and rotation) to symmetric positive definite matrices (SPD), aligning the source and target domains to the same subspace. Although this method can be applied across subjects and sessions, its practicality is limited due to the output being SPD matrices. Shared Latent Response (SLR) \cite{masaki2020facil} uses common spatial filtering methods, including CCA and TRCA, to extract features from the training data and then uses least squares regression to obtain new spatial filters that project test data onto the same subspace as the training data. This approach is applicable to cross-subject, cross-session, and cross-device scenarios, with input data being common time series data, providing more flexibility in practical applications. ALign and Pool for EEG Headset domain Adaptation (ALPHA) \cite{liu2022alpha} aligns spatial patterns through orthogonal transformations and aligns the covariance between different distributions using linear transformations, mitigating variations in spatial patterns and covariance. This method further improves upon obtaining new spatial filters and achieves better performance than SLR. Tangent Space Alignment (TSA) \cite{bleuze2022tsa} shares similarities with RPA as it aligns different domains to the same subspace through translation, scaling, and rotation, but operates within the tangent space. The tangent space being Euclidean allows for faster decoding, and rotation can be achieved with a singular value decomposition (SVD), making it computationally efficient compared to RPA. The tangent vector, as the output of TSA, has limited compatibility to most classification methods for SSVEP detection that requires time-series signals and thus its practicality is restricted. Similar issues are also found in applying subspace alignment methods that identify target stimuli by computing correlation coefficients between spatial features in the same subspace \cite{masaki2020facil, liu2022alpha}.

\subsection{Data Alignment}
Data alignment methods perform alignment between source domain samples and target domain samples in order to mitigate the disparities between the two domains. Recently, the Least Squares Transformation (LST) approach \cite{Chiang2021LST} finds a linear transformation relationship among the SSVEP data, effectively reducing the errors between the transformed data from the source SSVEP and the target SSVEP. This method is applicable in multiple cross-domain scenarios, including cross-sessions, cross-subjects, and cross-devices transfer learning for boosting the calibration of training-based SSVEP detection. Furthermore, the property of using time-series output allows data alignment methods to be seamlessly integrated into common SSVEP detection methods such as TRCA \cite{nakan2018trca}, and Conv-CA \cite{li2020conCCA}. The utilization of data alignment for SSVEP-based BCI can significantly enhance their feasibility in real-world applications with reduced calibration effort for individual users. Yet, as LST has high flexibility for cross-domain adaptation and desirable simplicity, it can only transform SSVEP signals within the same visual stimulus via linear combination. Further development of data alignment techniques for SSVEP signals is required to tackle the issues of stimulus-independent training and non-linear transformation learning.

\section{Methodology}
\subsection{SSVEP-DAN}
In this study, we assume that there exists a non-linear and channel-wise transformation of SSVEP signals between subjects. We propose a neural network-based transformation to transfer SSVEP signals from an existing subject (source domain) to a set of additional calibration data for a target subject (target domain). The architecture of the proposed neural network is illustrated as in Figure \ref{fig:Network}. 
The input to SSVEP-DAN is SSVEP data obtained in the source domain sizes  $\mathbb{R} ^{N_C \times N_S}$, while the output matches the average of multiple trials corresponding to specific stimuli from the target subject, formatted as $\mathbb{R} ^{N_C' \times N_S}$. Here, $N_C$ denotes the number of input channels, $N_{C'}$ is the number of output channels, and $N_S$ represents the number of sampling points. We employ these input and output sets to train our SSVEP-DAN model.

Conventional SSVEP detection methods, such as CCA and TRCA, apply spatial filtering to find a linear combination of channel-wise SSVEP signal \cite{lin2007cca,nakan2018trca}. These methods have been demonstrated to improve signal-to-noise ratio (SNR) \cite{Johnson2006snr} and enhance SSVEP detection performance. Additionally, SSVEP features time-synchronous signals with prominent oscillatory waveform at specific stimulation frequency and its harmonics, we assume the cross-domain adaptation requires transformation in spatial domain rather than in temporal domain. The signal processing of spatial filtering has been utilized in recently developed neural network-based EEG decoders such as SCCNet \cite{wei2019sccnet}, where a spatial convolutional layer serves for noise reduction and feature extraction. Therefore, in the first module, we utilize spatial convolution, incorporating $N_{F}$ spatial filters with a shape of $(N_C,1)$, to project the original SSVEP data into latent spaces to obtain spatial features, where $N_{F}$ represents the number of spatial filters. Note that in this study, the number of spatial filters, $N_{F}$, is equal to the number of input channels, $N_{C}$. 
After the convolutional layer, the dimensions of the latent features are properly permuted and subjected to batch normalization, where each timestamp and filter channel is normalized independently. Subsequently, channel-wise fully connected layers with dimensions $\mathbb{R} ^{N_C' \times N_S}$ are applied, projecting the data from spatial filter component spaces to output channel spaces.

To capture non-linear channel relations between spatial features and target SSVEP templates, we incorporate two channel-wise fully connected layers with a hyperbolic tangent activation function in between. Channel information at each time point is integrated using another channel-wise fully connected layer, projecting the feature into a new latent space. The activation function $tanh$ is then applied to facilitate model fitting. Lastly, another channel-wise fully connected layer is used to project the features into the target domain space, and permutation is applied to obtain transformed SSVEP data in the target domain.

\subsection{Training Strategies}
\subsubsection{Stimulus-independent training}
Based on the assumption that the transformation of SSVEP data between subjects is irrelevant to the visual stimuli, we propose a stimulus-independent training strategy dedicated for neural network-based SSVEP data alignment to augment the data amount through combining data across different stimuli, aiming for a relatively robust transformation under limited data amount.
The stimulus-independent training process involves cross-stimulus training that enables the learning of model parameters across different stimulus frequencies and thereby ensures a relatively reliable learning process for the transformation between subjects, particularly when calibration data is scarce.
As data of different stimulus types are merged within individual training batch, the stimulus type between the source/target SSVEP data remains the same. 

\subsubsection{Two-phase model training}
\begin{figure}[!ht]
    \centering
    \includegraphics[width=\columnwidth, ]{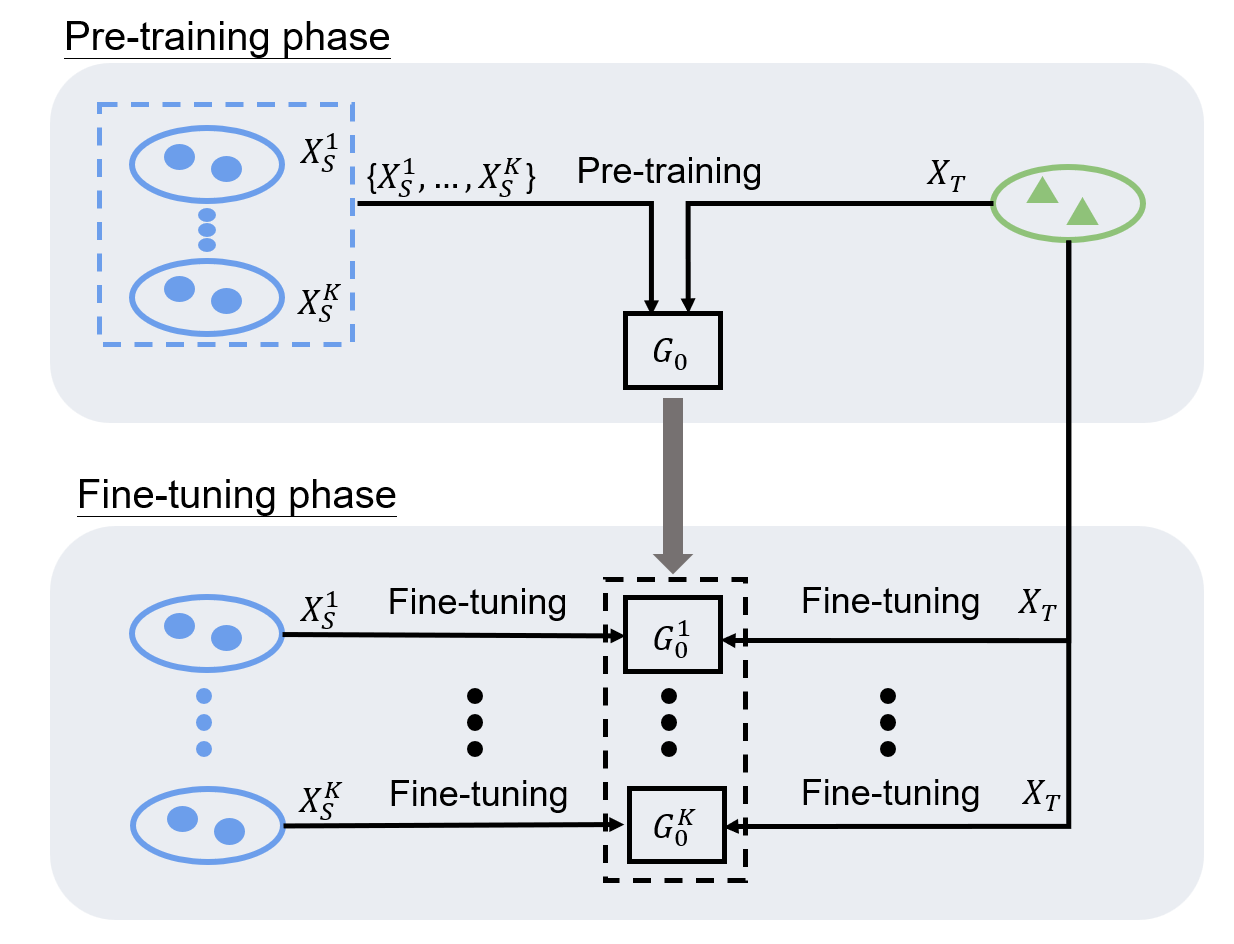}
    \caption{
    The two-phase model training proposed in this study includes a pre-training phase and a fine-tuning phase. Initially, a pre-trained model, denoted as $G_0$, is trained using the target data $X_T$ and the entire source data pool, represented as ${X^1_S, X^2_S, ..., X^K_S}$, where $K$ is the total number of source domains. Subsequently, the pre-trained model $G_0$ undergoes separate fine-tuning with each individual source dataset, resulting in fine-tuning models $G^1_0, G^2_0, ..., G^K_0$.
    }
    \label{fig:two_phase}
\end{figure}

\begin{figure}[!ht]
    \centering
    \includegraphics[width=\columnwidth, ]{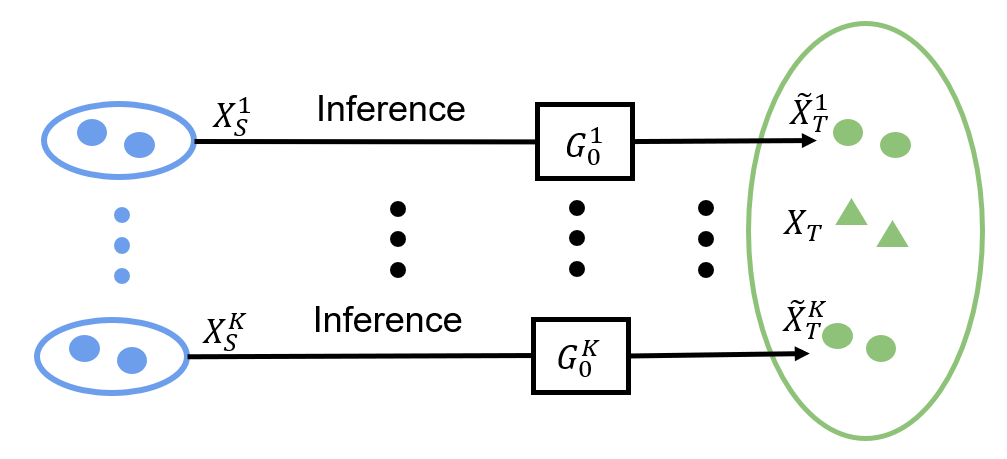}
    \caption{
    The inference procedure transforms the data of each source domain into additional data in the target domain. For each source domain $D^k_S$, $\forall k\in\{1, 2, ..., K\}$, the SSVEP-DAN model $G^k_0$ transforms the source data $X^k_S$ into $\widetilde{X}^k_T$ for supplement the amount of data in the target domain $D_T$.
    }
    \label{fig:inference_phase}
\end{figure}

Recent studies of deep-learning-based SSVEP detection suggest that fine-tuning a pre-trained model improves the performance as the training data can be fully used during pre-training and the characteristic of individual domain is considered during fine-tuning \cite{wei2019sccnet, ravi2020ssvepIDUD, guney2021ssvepDNN}.
We adopt this strategy to handle the training of SSVEP-DAN in the case of multiple source domains with a two-phase procedure consisting of a pre-training phase and a fine-tuning phase, as illustrated in Figure \ref{fig:two_phase}. In the pre-training phase, data from all source domains are concatenated and used in the training of a pre-trained model. Next, a fine-tuning phase is performed to fine-tune the pre-trained model separately using the data from individual source domain and acquire a fine-tuned models between each source domain and the target domain. Lastly, all transformed data from the source domains merge into the calibration data for the target subject, as depicted in Figure \ref{fig:inference_phase}.

\subsection{Model Fitting}
The network is trained using the Adam optimizer \cite{kingma2014adam} with a learning rate set to ${5\times10^4}$. The transformation loss $L_{trans}$ is determined by the Mean Square Error (MSE) between the target SSVEP data $X_T$ and the output SSVEP data $\widetilde{X}_T$: 
\begin{equation}
    L_{trans}(X_T, X'_T) = \frac{1}{N}\sum_{i=1}^{N}||X_T - \widetilde{X}_T||^2_2,
\end{equation}
where $N$ is the batch size. The target SSVEP data $X_T$ is obtained by averaging calibration trials of the same stimulus from the target subject as suggested in \cite{Chiang2021LST}.
During the pre-training phase, the source subjects are divided into a training set and a validation set using a subject-wise ratio of 8:2. The model is trained for 500 epochs, and the model weights that yield the lowest validation set loss are employed as the fitting result. In the fine-tuning phase, a 150-epoch fitting process is conducted using data from a single source subject, following the same configuration as described above, except for the training/validation data splitting.

\subsection{TRCA-Based Performance Assessment}
TRCA is a training-based algorithm that aims to extract task-related components by maximizing the inter-trial coherence of neural activity across multiple trials within each specific task \cite{nakan2018trca}. Furthermore, the combination of TRCA and filter bank analysis facilitates the decomposition of SSVEP signals into multiple sub-band components, thereby effectively extracting independent information embedded within the harmonic components \cite{chen2015fbcca}. Finally, an ensemble approach is employed to integrate multiple filters trained using the aforementioned methods.

\subsection{Data}
\subsubsection{Dataset I: SSVEP Benchmark Dataset}
\cite{wang2017benchmark} used in this study is a publicly available SSVEP dataset prepared by the Tsinghua group. In this dataset, the SSVEP-based BCI experiment involved 35 participants. Each participant participated in 6 blocks of the experiment, each block comprising 40 trials presented in random order. Visual stimuli were presented within a frequency range of 8 to 15.8 Hz with an interval of 0.2 Hz. The phase values of the stimuli ranged from 0, with a phase interval of 0.5$\pi$. EEG signals were collected utilizing the Synamps2 EEG system (Neuroscan, Inc.). They were recorded using the extended 10-20 system through 64 channels. We selected EEG data from eight channels (PO3, PO4, PO5, PO6, POz, O1, O2, Oz) in the analysis and performance evaluation. The EEG signals were down-sampled from 1000 to 250 Hz, and a notch filter at 50 Hz was applied to remove the common power-line noise. The data were extracted in [$L_1$ s, $L_1$ + $T_{w_1}$ s] of the stimulus onset, where $L_1$ is the latency delay ($L_1 = $ 0.14 s) and $T_{w_1}$ indicates the time-window length ($T_{w_1} = $ 1.5 s). In the SSVEP detection based on filter-bank TRCA, we set the number of filter banks as five for this dataset as previously suggested \cite{li2020conCCA}.

\subsubsection{Dataset II: Wearable SSVEP BCI Dataset}
The Wearable SSVEP BCI dataset \cite{zhu2021wearalbe} used in this study is another publicly available SSVEP dataset released by the Tsinghua group.
In this dataset, 102 healthy subjects participated in the wearable SSVEP-based BCI experiment. The experiment consisted of 10 blocks, each of which contained 12 trials in random order of 12 visual stimuli. Stimulation frequencies ranged from 9.25 to 14.75 Hz with an interval of 0.5 Hz. The phase values of the stimuli started at 0, and the phase difference between two adjacent frequencies was 0.5$\pi$. EEG signals were collected utilizing the Neuracle EEG Recorder NeuSen W (Neuracle, Ltd.) system. The 8-channel EEG data was recorded using wet and dry electrodes, and the electrodes were placed according to the international system 10-20. All channels (PO3, PO4, PO5, PO6, POz, O1, O2, Oz) of the EEG signals were used in data analysis and performance evaluation. The EEG signals were resampled at 250 Hz from 1000 Hz. To remove the common power-line noise, a 50 Hz notch filter was applied to the dataset. The data was extracted in [0.5 + $L_2$ s, 0.5 + $L_2$ + $T_{w_2}$ s], where 0.5 s denotes stimulus onset, $L_2$ indicates latency delay ($L_2 $ = 0.14 s) and $T_{w_2}$ is the time-window length ($T_{w_2} = $ 1.5 s). For this dataset,  we set the number of filter banks as three according to \cite{zhu2021wearalbe}.

\subsection{Performance Evaluation}

The performance evaluation of the SSVEP-DAN was conducted through leave-one-subject-out cross-validation. In this approach, each subject is treated as the target subject, while the remaining subjects serve as source subjects, providing existing SSVEP data. This setup reflects the real-world scenario of employing SSVEP-based BCI when introducing a new user to the system.

In the leave-one-subject-out cross-validation, we designated a specific set of trials as the test data for each individual target subject. Within each subject's trials from Dataset I, the initial 4 trials were utilized as both the calibration data and the target SSVEP data for the training of data alignment, leaving the remaining 2 trials as the designated test data. For Dataset II, the first 6 trials were allocated for calibration or data alignment training, with the remaining 4 trials serving as the designated test data. We followed the precedent set by a prior study \cite{Chiang2021LST} by using a minimum of 2 calibration trials. This minimum is necessitated by the TRCA, which requires at least 2 calibration trials for the acquisition of an average SSVEP template.

To justify the efficacy of this framework, we conducted a performance comparison of SSVEP-DAN against the Baseline (without domain adaptation) and other domain adaptation methods, specifically Concatenation (Concat.) and LST. The performance comparison was based on the evaluation of SSVEP detection performance using filter-bank TRCA with calibration data prepared using these schemes, considering different numbers of calibration trials (Dataset I: 2-4; Dataset II: 2-6) from the target (test) subject. A detailed description of these schemes is provided below.

\subsubsection{Baseline}
This approach utilizes the filter-bank TRCA without any provided source data. The TRCA relies on calibration data collected solely from the target subject.

\subsubsection{Concat.}
The Concat. approach employs a simple transfer learning scheme by naively concatenating all SSVEP data from the source domains with the target SSVEP data without any transformation. The concatenated SSVEP data is then used as calibration data for TRCA.

\subsubsection{LST}
The LST approach involves a linear transformation to transfer the source SSVEP data to the target domain based on source subjects and stimuli. The LST-transformed data is then concatenated with the target SSVEP data to form the calibration data for TRCA.

\subsubsection{SSVEP-DAN}
The SSVEP-DAN approach utilizes a non-linear transformation to transfer the source SSVEP data to the target domain based on source subjects. The transformed data is then concatenated with the target SSVEP data to construct the calibration data for TRCA.

In addition to the Baseline scheme, for Concat. , LST, and SSVEP-DAN schemes, the training and validation sets were randomly partitioned. Furthermore, the parameters of the SSVEP-DAN model were initialized randomly. The decoding performance of each domain adaptation scheme was estimated by averaging ten repeats. We utilized the Wilcoxon signed-rank test to assess the statistical significance of the improvements between the proposed SSVEP-DAN-based method and other domain adaptation schemes.

\begin{table}[h!]
 \caption{Domain Adaptation Tasks}

\centering
\resizebox{0.45\textwidth}{!}{
\begin{tabular}{lll}
\toprule
\textbf{Task} &\textbf{Source domain} &\textbf{Target domain} \\
\midrule
Benchmark     & Dataset I & Dataset I \\
Dry to dry   & Dataset II - Dry & Dataset II - Dry\\
Wet to wet   & Dataset II - Wet & Dataset II - Wet\\
Dry to wet & Dataset II - Dry & Dataset II - Wet \\
Wet to dry     & Dataset II - Wet & Dataset II - Dry \\
\bottomrule
\end{tabular}}
  \label{tab:cases_domain}
\end{table}
\section{Results and Discussion}

This section depicts a series of experimental results with interpretations of our findings that demonstrate the traits and efficacy of the data-alignment domain adaptation based on the proposed SSVEP-DAN. We performed an in-depth evaluation in the context of scenarios in practical SSVEP-based BCI applications. Using the two datasets, we conducted a total of five domain adaptation tasks where the proposed SSVEP-DAN as shown in Table \ref{tab:cases_domain}. For all adaptation schemes, the performance was compared against 1) the number of calibration trials obtained from a target subject, and 2) the number of source subjects. In addition, we present an ablation study as well as visualization of the adaptation results to reveal the intriguing features of the SSVEP-DAN and the effect of training strategies.

\begin{figure*}[!]
\centering
    \subfloat[]{\includegraphics[width=.33\linewidth,trim={3.8cm 0 4cm 0},clip]{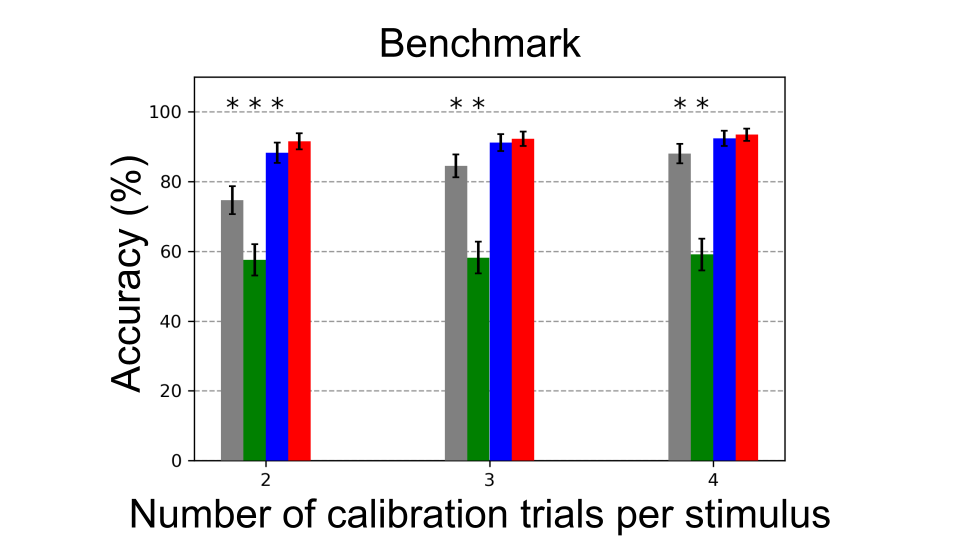}}\hfill
    \subfloat[]{\includegraphics[width=.33\textwidth,trim={3.8cm 0 4cm 0},clip]{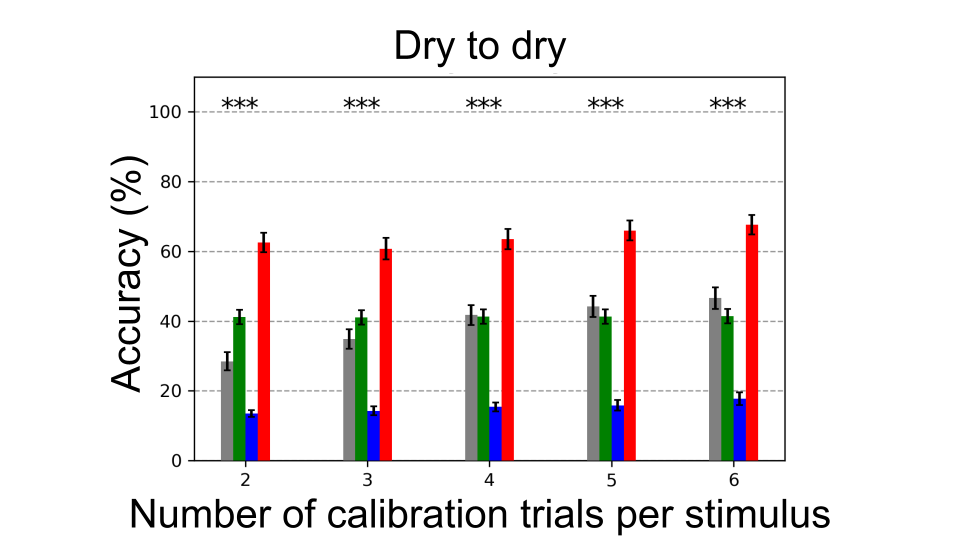}}\hfill
    \subfloat[]{\includegraphics[width=.33\textwidth,trim={3.8cm 0 
    4cm 0},clip]{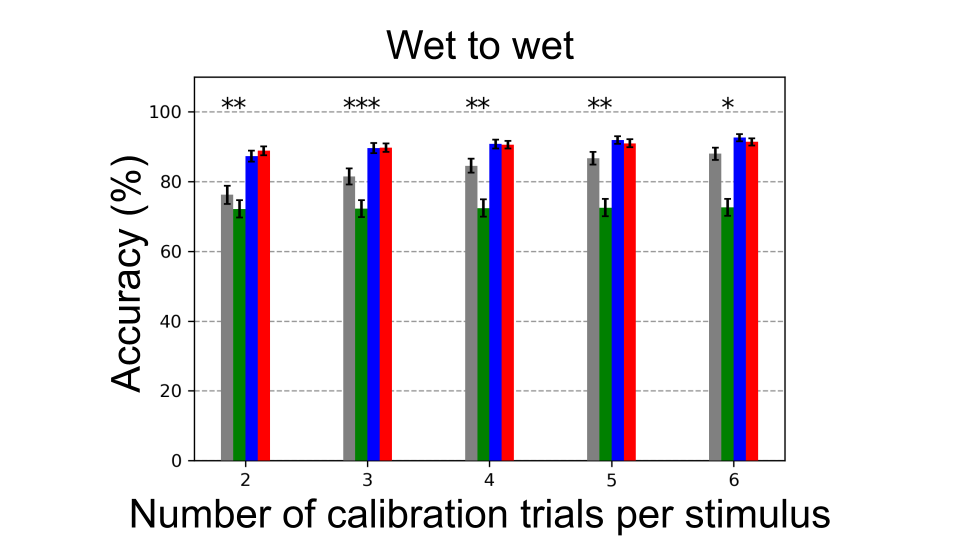}}\hfill
    
    \subfloat[]{\includegraphics[width=0.33\textwidth,trim={3.8cm 0 4cm 0},clip]{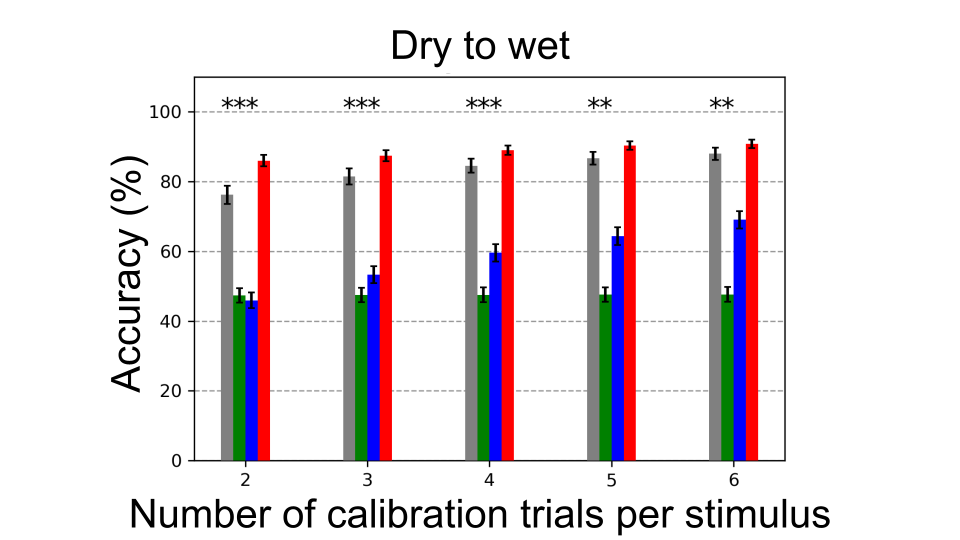}}
    \subfloat[]{\includegraphics[width=0.33\textwidth,trim={3.8cm 0 4cm 0},clip]{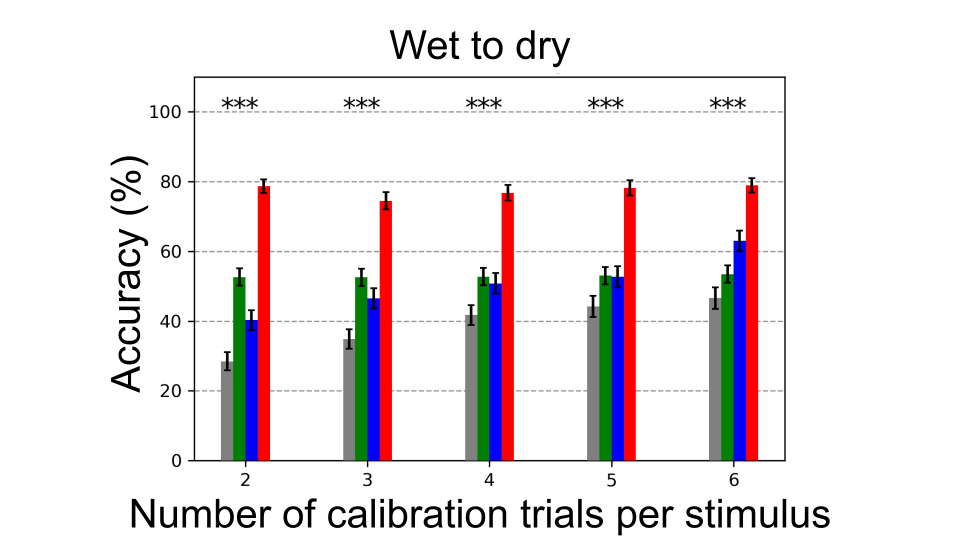}}
    \subfloat{\includegraphics[width=0.33\linewidth,trim={3.8cm 0 4cm 0},clip]{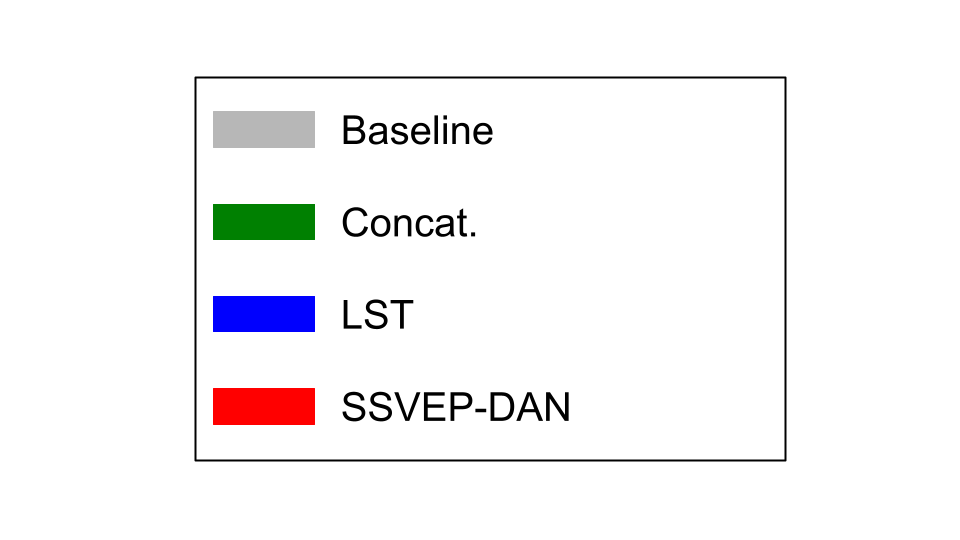}}
\caption{The performance evaluation ($\%$) against number of calibration trials per stimulus in the target domain across the five scenarios. Asterisks indicate significant differences between SSVEP-DAN and other methods. (*$p<$0.05).}
\label{fig:tps}
\end{figure*}

\subsection{Performance Comparison}
Figure \ref{fig:tps} depicts the cross-domain performance using different schemes under varying numbers of calibration trials for each stimulus. The results indicate that the proposed SSVEP-DAN significantly outperforms the other three schemes in most cases, regardless of the number of calibration trials from the target subject. 
Moreover, as the number of calibration trials for the target subject increases, our proposed method consistently improves the performance of the SSVEP decoding algorithm. 
Interestingly, although in certain scenarios, such as 'benchmark', 'dry to wet', or 'wet to wet' scenarios, the naive transfer learning method (Concat. scheme) appears to have a negative impact on TRCA-based methods, it exhibits a positive influence on TRCA-based methods when dry electrode devices are used as recording equipment for new subjects.
One possible explanation for this observation is that data variability and data quality may concurrently influence the performance of training-based detection methods. When the data quality of new users is high, the impact of data variability becomes more prominent, as high-quality data may be more sensitive to subtle variations. Conversely, when the data quality of new users is relatively low (i.e. dry-electrode EEG data), there is a greater demand for higher-quality data, and simultaneously, the influence of data variability becomes less significant. 
Therefore, when training-based detection methods can accurately classify using high-quality data from new users, the negative impact of the Concat. scheme is exacerbated due to the heightened influence of data variability. In contrast, when training-based detection methods struggle to accurately classify using low-quality data from new users, acquiring a larger dataset of higher quality mitigates the negative impact of the Concat. scheme.
These differences in signal quality become particularly evident when comparing dataset II to the benchmark dataset I \cite{zhang2022bidirectional, bassi2022fbdnn}. Furthermore, the SNR of dry-electrode data is typically lower than that of wet-electrode data due to factors such as unstable contact, sensitivity to artifacts, or higher impedance \cite{mihajlovic2013dry, xing2018high}. Within dataset II, signal quality is observed to be lower under dry-electrode data compared to wet-electrode data \cite{zhu2021wearalbe}.
On the other hand, although LST-based methods can effectively enhance TRCA-based methods in some scenarios, including 'benchmark', 'wet to dry', and 'wet to wet' scenarios, LST-based methods also have a detrimental effect on TRCA-based methods and lead to negative transfer \cite{pan2010survey, zhang2022survey} when the source data is collected through dry electrode devices. 
This implies the stability of the transformation matrix in LST is highly susceptible to the quality of SSVEP signals \cite{Chiang2021LST}. Since trial averaging can enhance the signal-to-noise ratio (SNR) of SSVEP \cite{nakan2018trca, nakanishi2014high}, using this signal as the output can reduce the impact of signal quality. However, when the quality of the input SSVEP signal is poor, it may result in less reliable transformation matrices, consequently affecting the final transformation performance. Hence, we find that LST can effectively improve TRCA-based methods when data from source subjects are of high quality. Conversely, when data from source subjects are of lower quality, LST can lead to negative transfer. Furthermore, it is observed that increasing the number of calibration trials for each stimulus in most scenarios enhances the performance of LST, as it allows for obtaining better output SSVEP signal. 
It is worth noting that our proposed approach consistently demonstrates significant improvements over TRCA-based methods across various scenarios, especially when calibration data is scarce. This improvement is attributed to SSVEP-DAN's capability to transform source domain data into target domain, effectively expanding the training dataset for training-based decoding algorithms and subsequently enhancing decoding performance. In particular, the results characterize the capability of the proposed SSVEP-DAN in real-world practical applications where dry-electrode SSVEP data were involved \cite{xing2018high}.

\begin{figure*}[!]
\centering
    \subfloat[]{\includegraphics[width=.33\textwidth,trim={3.8cm 0 4cm 0},clip]{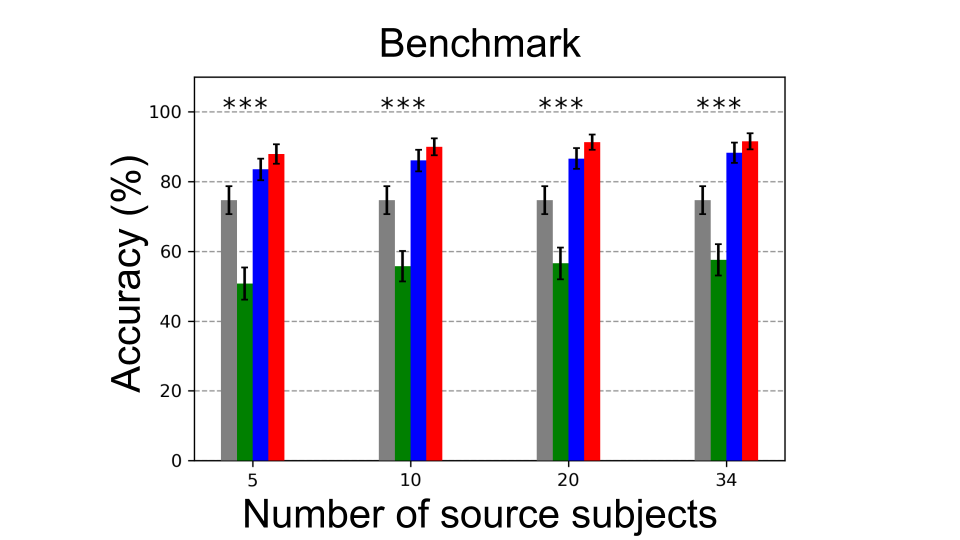}}\hfill
    \subfloat[]{\includegraphics[width=.33\textwidth,trim={3.8cm 0 4cm 0},clip]{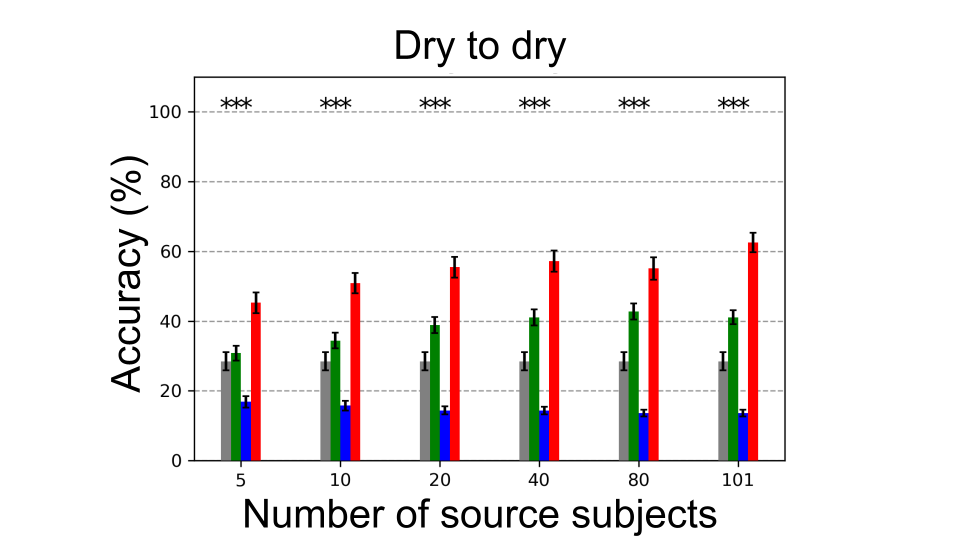}}\hfill
    \subfloat[]{\includegraphics[width=.33\textwidth,trim={3.8cm 0 4cm 0},clip]{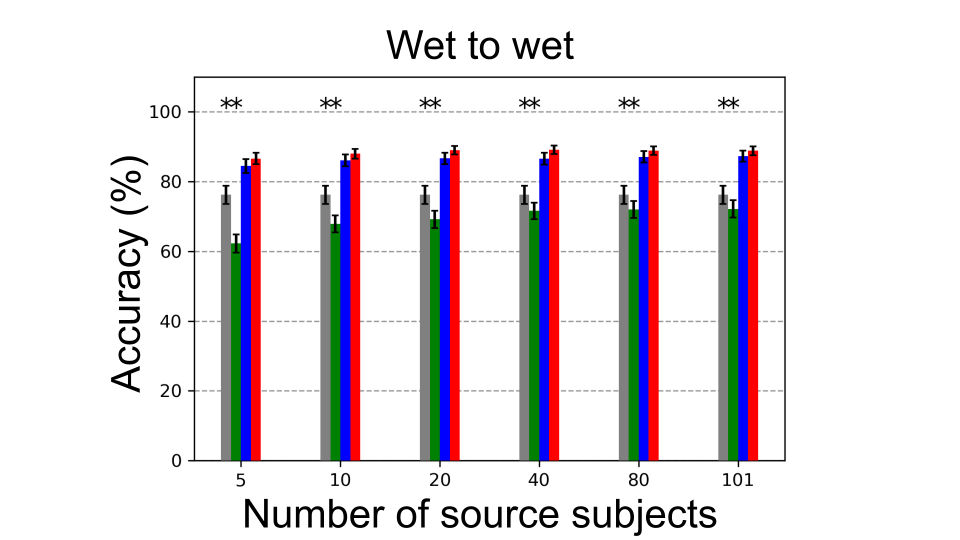}}\hfill
    
    \subfloat[]{\includegraphics[width=0.33\textwidth,trim={3.8cm 0 4cm 0},clip]{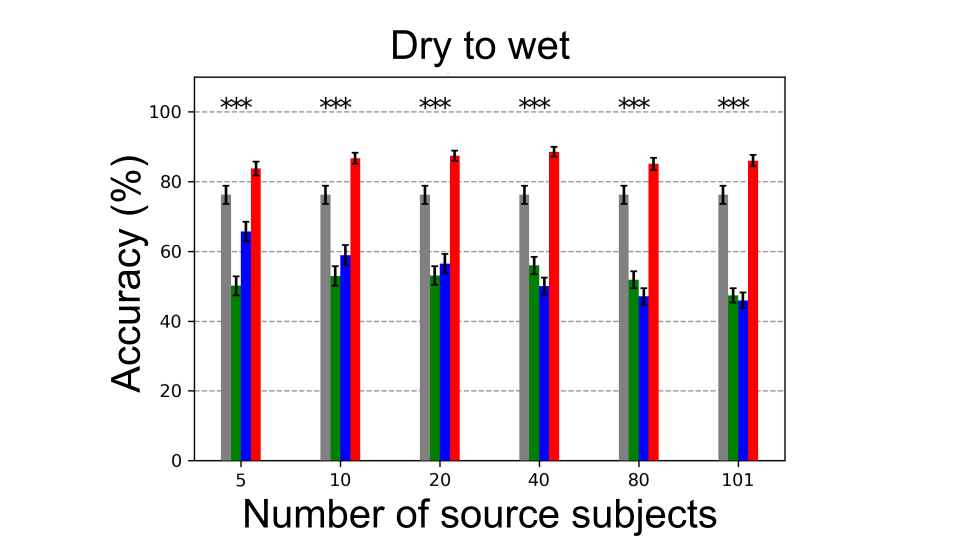}}
    \hspace*{0.005\textwidth}%
    \subfloat[]{\includegraphics[width=0.33\textwidth,trim={3.8cm 0 4cm 0},clip]{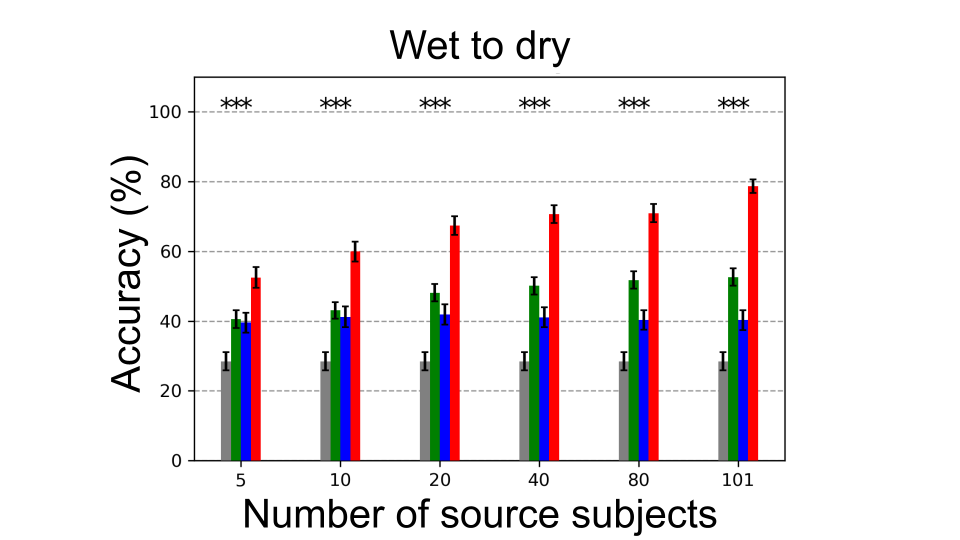}}
    \subfloat{\includegraphics[width=0.33\textwidth,trim={3.8cm 0 4cm 0},clip]{figures/performance_legend_v7.png}}

\caption{
The performance evaluation ($\%$) against number of source subjects across the five scenarios. A fixed amount of target data was applied in the analysis (two trials per stimulus). Asterisks indicate significant differences between SSVEP-DAN and other schemes. (*$p<$0.05).}
\label{fig:supp}
\end{figure*}

\subsection{Number of Source Subject}
This section validates the performance of the SSVEP-DAN against the numbers of source subjects. 
For the Concat., LST, and SSVEP-DAN schemes, a subset of source subjects was randomly selected from all of the source subjects with ten repetitions.
In both Dataset I and Dataset II, we assess the performance in the case of insufficient number of source subjects, with a fixed amount of target data (two trials). 
Figure \ref{fig:supp} illustrates the cross-domain performance against the number of source subjects. The results demonstrate the superiority of the proposed SSVEP-DAN regardless of the number of source subjects in most cases. 
The effect of source subjects is relatively visible in the tasks of 'dry to dry' and 'wet to dry' where training-based detection methods relies on supplement of transformed SSVEP data to handle the low quality of the calibration data.
Meanwhile, the observation of negative transfer by the Concat. and the LST is analogous to that in Fig. \ref{fig:tps}. 
Additionally, we observed that when high-quality source data from the source domain increases, the LST-based methods slightly improve, particularly in scenarios like the 'benchmark', 'wet to wet', and 'wet to dry'. However, when low-quality source data from the source domain increases, the negative impact of LST on TRCA-based methods intensifies, particularly in 'dry to dry' and 'dry to wet' scenarios. The proposed SSVEP-DAN, trained on a large amount of data, is capable of improving the performance of SSVEP detection consistently across number of source subjects.

\subsection{Ablation Study}
\begin{table}[b!]
 \caption{Ablation analysis on the efficacy of the key elements in the proposed SSVEP-DAN framework. The asterisks indicate a significant difference between the proposed method, its reduced variation, and the baseline. (*$p<$0.05, **$p<$0.01, ***$p<$0.001).}

\centering
\resizebox{0.49\textwidth}{!}{
\begin{tabular}{llllll}
\toprule
\textbf{Methods} &\textbf{Benchmark} &\textbf{Dry to dry} & \textbf{Wet to wet} &\textbf{Dry to wet} &\textbf{Wet to dry}\\
\midrule
Proposed     & 91.56 & 62.55 & 88.67 & 86.01 & 78.67 \\
\midrule
(w/o) stimulus-independent training  & 84.79\textsuperscript{***} & 39.09\textsuperscript{***} & 87.72 & 81.34\textsuperscript{**} & 63.02\textsuperscript{***} \\ 

(w/o) pre-training phase & 90.60 & 53.43\textsuperscript{***} & 88.28\textsuperscript{***} & 82.67\textsuperscript{***} & 69.65\textsuperscript{***}\\ 

(w/o) fine-tuning phase     & 89.15\textsuperscript{***} & 34.55\textsuperscript{***} & 85.24\textsuperscript{***} & 75.15\textsuperscript{***} & 52.67\textsuperscript{***}\\ 
\midrule
(w/o) non-linear activation function & 91.31 & 44.71\textsuperscript{***} & 87.49\textsuperscript{*} & 84.75\textsuperscript{*} & 69.02\textsuperscript{***} \\

(w/) temporal convolution  & 58.03\textsuperscript{***} & 28.79\textsuperscript{***} & 38.19\textsuperscript{***} & 38.35\textsuperscript{**} & 25.48\textsuperscript{***} \\
\midrule
Baseline (w/o adaptation)  & 74.68\textsuperscript{***} & 28.45\textsuperscript{***} & 76.21\textsuperscript{***} & --- & ---  \\
\bottomrule
\end{tabular}}
  \label{tab:ablation}
\end{table}
We conducted an ablation analysis to evaluate the effectiveness of main training strategies of SSVEP-DAN, namely, stimulus-independent training and a two-phase training approach, as well as the components in the model architecture. Table \ref{tab:ablation} presents the results of SSVEP-DAN using different training methods and different model configurations, with reference to the proposed method and the baseline (no adaptation).
In the 'w/o stimulus-independent training' approach,  the SSVEP-DAN was trained to align data of each stimulus specifically.
In the 'w/o pre-training phase' approach, for each source domain subject, we skipped the pre-training phase and trained the SSVEP-DAN using source data individually in the fine-tuning phase.
In the 'w/o fine-tuning phase' approach, we utilized data from multiple source domains to train the model to obtain the pre-training model and without the subsequent fine-tuning on each source domain.
For validating the efficacy of components in the proposed network architecture, we conducted the 'w/o non-linear activation function' method that reduce the non-linearity introduced by the hyperbolic tangent activation layer. In addition, we conducted 'w/ temporal convolution' method to incorporate an additional temporal convolutional layer, following by the original spatial convolutional layer, to increase the complexity of the model.

First, in the majority of cases for both Dataset I and Dataset II, the complete SSVEP-DAN significantly outperforms the training method without stimulus-independent training. This is primarily attributed to cross-stimulus training, which enhances the available training data by combining data from different stimuli during training, resulting in a more robust model, particularly when calibration data is scarce. 
Furthermore, we observe that the performance of our proposed training method consistently surpasses that of each single-stage training method (without pre-training or without fine-tuning) in most cases for both Dataset I and Dataset II. This indicates that pre-training enables SSVEP-DAN to leverage data from multiple source domains to train the model, enhancing its ability to capture common features from SSVEP data and improve generalization, while avoiding overfitting to specific subjects. Moreover, given that individual subjects may exhibit unique neural responses, fine-tuning the pre-trained model based on the individual subject's characteristics allows it to adapt to the representations specific to the source-target subject pairs. In summary, these two primary training methods are indispensable in SSVEP-DAN as they provide data augmentation and robustness while enhancing the model's performance in aligning SSVEP data across different subjects and stimuli.

The result illustrates that our proposed model design consistently outperforms the design without a non-linear activation function across most scenarios in both Dataset I and Dataset II. This observed improvement can be attributed to several factors. 
The utilization of the hyperbolic tangent activation function is beneficial for training performance as it could mitigate signal drift or bias in the data by concentrating data near zero, offering an improvement in managing data variations \cite{nagarajan2021investigation, xu2023analysis}. Furthermore, it aids in the normalization and constraint of high-amplitude EEG data, thereby enhancing network stability \cite{nagarajan2021investigation}.
Regarding the complexity of the model architecture, we observed that incorporating the temporal convolutional layer deteriorates the performance in both Dataset I and Dataset II. This could be relevant to the excessive complexity of the model that leads to overfitting on the training data. This risk is particularly pronounced when the availability of EEG data is limited, making overly complex models susceptible to overfitting \cite{he2021data}.

\subsection{Visualization of Data Alignment}

\begin{figure}[!]
\centering
\subfloat[Benchmark]
{\includegraphics[width=.45\textwidth,trim={15cm 0 15cm  0},clip]{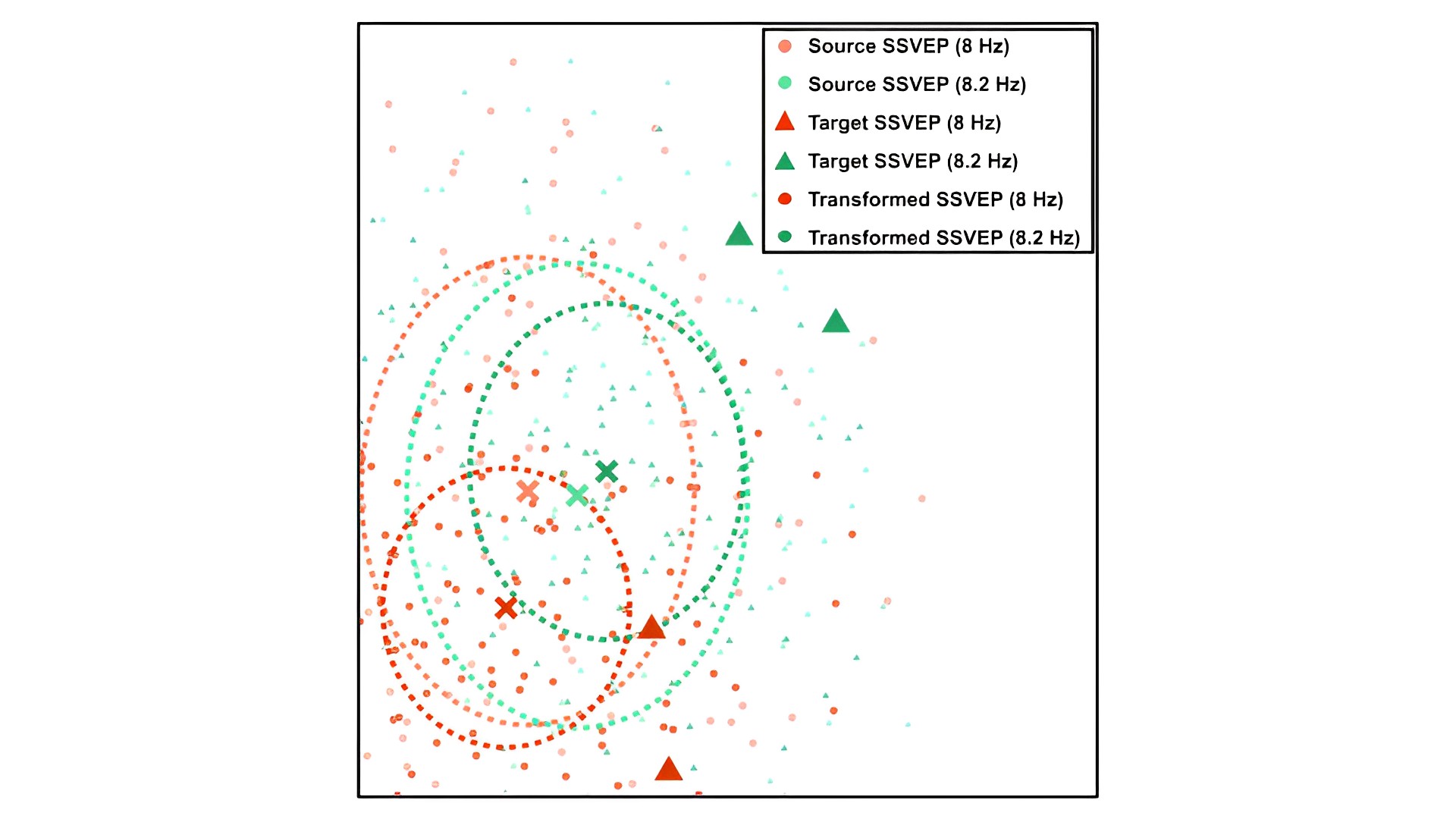}}\\
\subfloat[Wet to dry]
{\includegraphics[width=.45\textwidth,trim={15cm 0 15cm  0},clip]{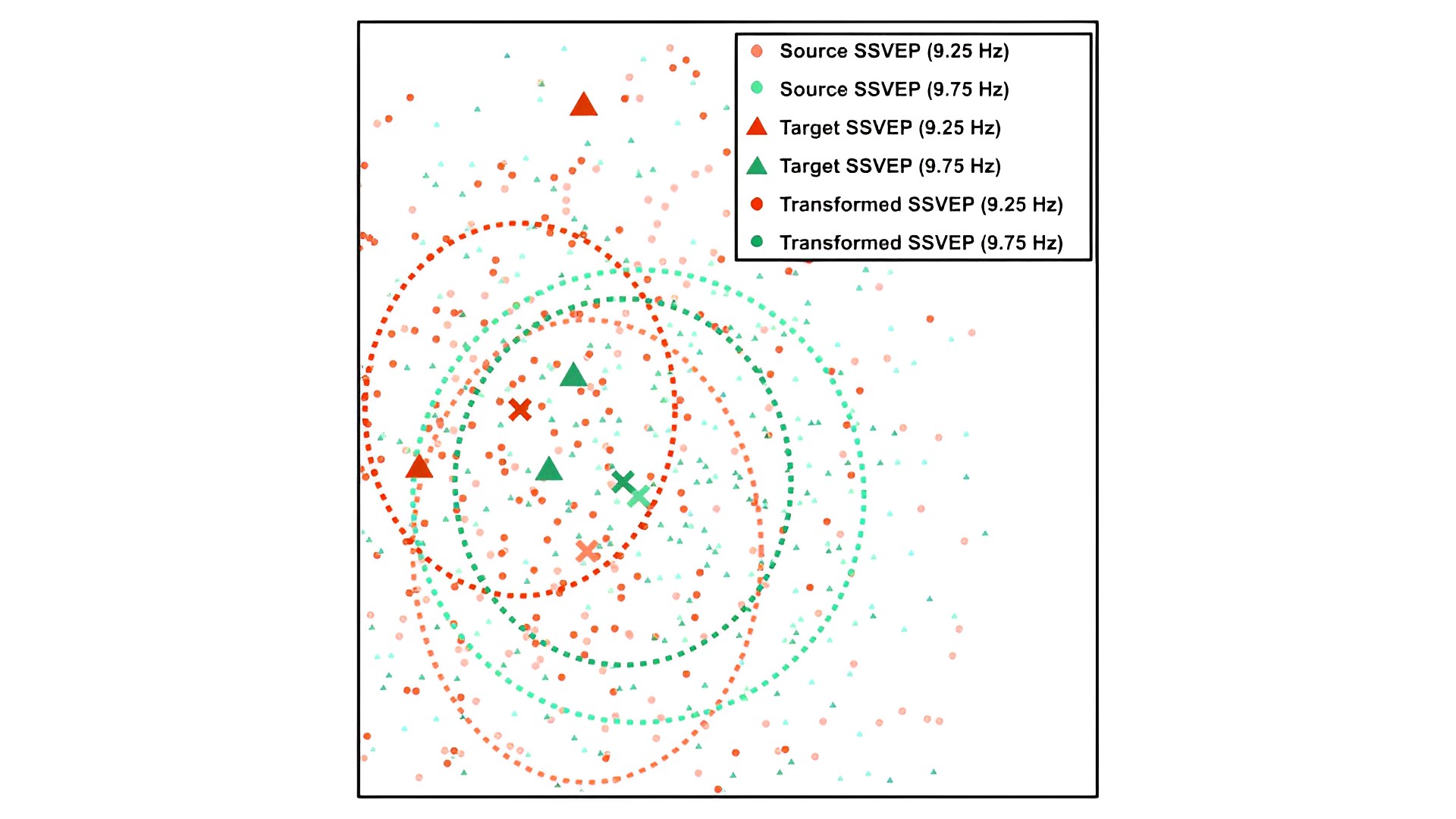}}
\caption{
Visualization of the source, target, and transformed SSVEP data based on t-SNE for the tasks (a) 'Benchmark' and (b) 'wet to dry'. We selected the SSVEP data for two stimuli ('benchmark': 8 and 8.2 Hz; 'wet to dry': 9.25 and 9.75 Hz), denoted by red and green dots, across subjects to illustrate the impact of SSVEP data alignment. The red/green triangles mark the target SSVEP data (two trials per stimulus). The light and dark shading distinguish the source SSVEP data from the corresponding transformed SSVEP data, with the class centroid indicated by a cross and the standard deviation delineated by a dotted circle. The shift in the SSVEP data clusters demonstrates the effect of data alignment performed by the proposed SSVEP-DAN.
}
\label{fig:filterBank}
\end{figure}

\begin{figure*}[!]
\centering
\subfloat[Benchmark]{\includegraphics[width=.5\textwidth,trim={0cm 0 1cm 0},clip]{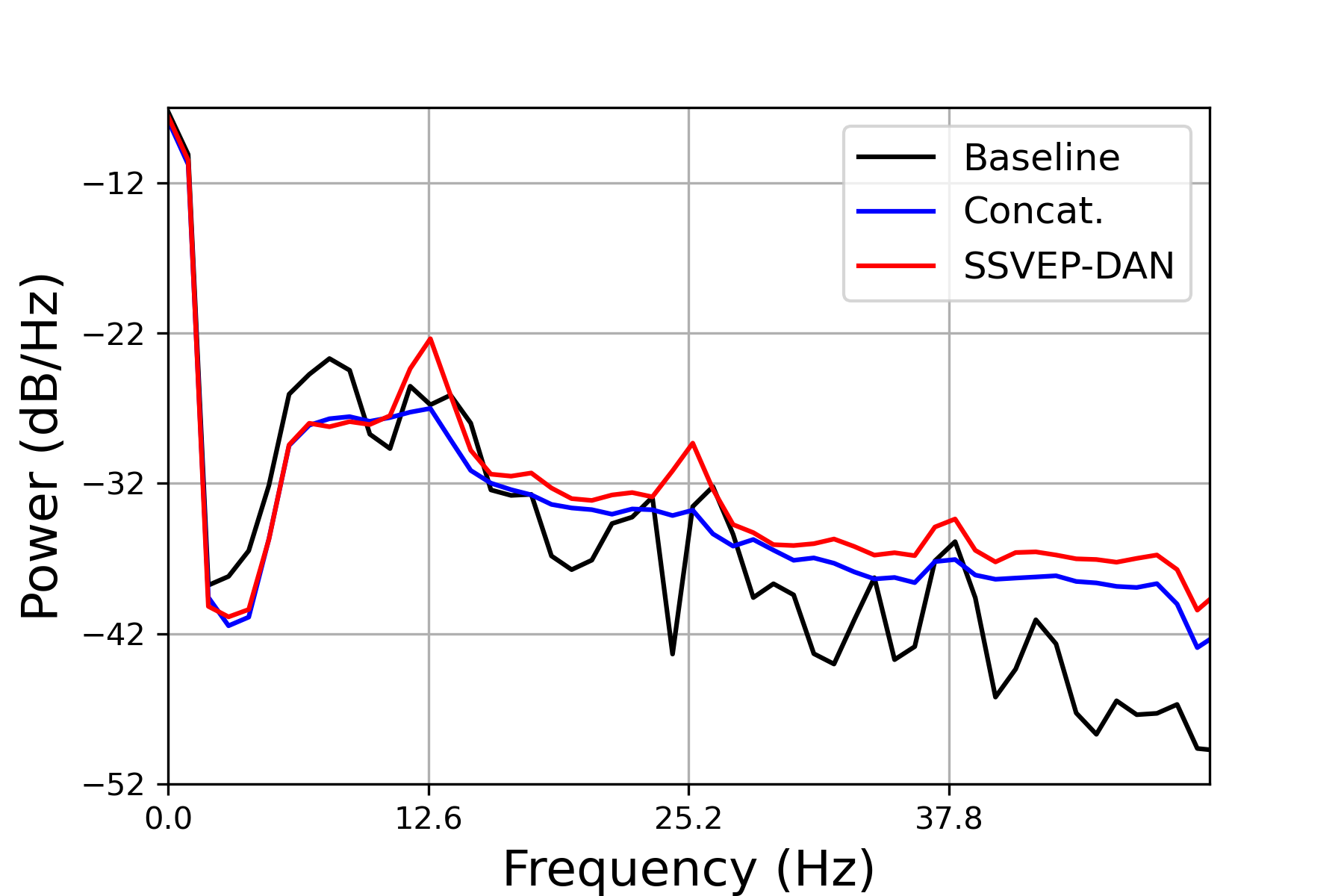}}\hfill
\subfloat[Wet to dry]{\includegraphics[width=.5\textwidth,trim={0cm 0 1cm 0},clip]{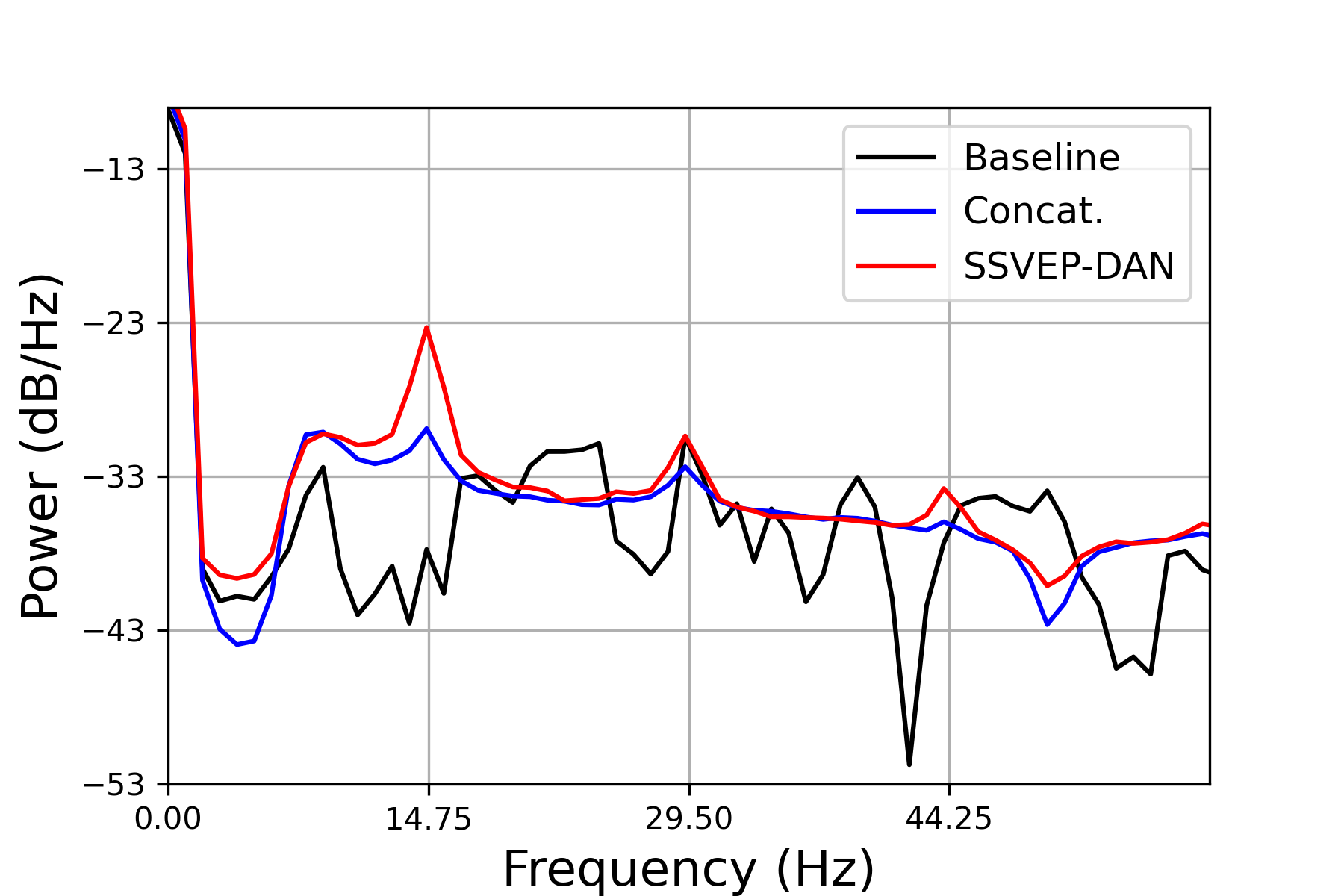}}\hfill
\caption{Averaged power spectrum density of the SSVEP data obtained under different conditions of domain adaptation in the (a) 'Benchmark' scenario with the 12.6 Hz stimulus and (b) 'wet to dry' scenario with the 14.75 stimulus. 'Baseline': average across two trials of target data only; 'Concat.': average across two trials of target data plus original source data; 'SSVEP-DAN': average across two trials of target data plus transformed data.}
\label{fig:psd}
\end{figure*}

We applied a statistical method, t-SNE \cite{van2008tSNE}, to embed the SSVEP data before and after applying alignment by SSVEP-DAN into two-dimensional scatters.
Figure \ref{fig:filterBank} (a) presents the t-SNE visualization of the subject S3 as the target subject of the Dataset I with all other subjects as the source subjects. Figure \ref{fig:filterBank} (b) illustrates the t-SNE visualization of the subject S1 as the target subject wearing a dry electrode device in the 'wet to dry' scenario of Dataset II. From these figures, we observed that under the same stimulus, the clusters of transformed SSVEP data were smaller compared to the clusters of source data. Additionally, the clusters of transformed data appear more separated. 
These findings suggest that the SSVEP-DAN method reduces inter-subject variability and increases inter-stimulus variability. It's worth noting that similar results were observed in a previous study \cite{Chiang2021LST}.
On the other hand, under the same stimulation, the transformed data exhibits a closer proximity to the target data compared to the source data, suggesting an enhanced similarity between the transformed and the target data.


Moreover, we utilized power spectrum analysis to investigate the effect of data alignment on the power spectral density (PSD) using the proposed SSVEP-DAN.
Figures \ref{fig:psd} (a) depict the PSD of the 12.6 Hz SSVEP signals for the subject S3 as the target subject in the Dataset I, under three schemes (Baseline, Concat., and SSVEP-DAN), when the calibration trials for each stimulus are two. Similarly, Figures \ref{fig:psd} (b) show the PSD of the 14.75 Hz SSVEP signals for the subject S1 as the target subject wearing a dry electrode device in the Wet-to-dry scenario from the Dataset II, with calibration trials two, for each stimulus under the three schemes. In the Concat. scheme, we observed different outcomes in Figure \ref{fig:psd} (a) and Figure \ref{fig:psd} (b). The results in Figure \ref{fig:psd} (a) indicate that the Concat. scheme on the Dataset I struggles to obtain stable spectra due to the high variability of SSVEP trials in the scheme, resulting in less concentrated spectral peaks at the target frequency. Conversely, the results in Figure \ref{fig:psd} (b) demonstrate that on the Dataset II, where the target participants have a suboptimal SNR, the Concat. scheme can slightly improve SNR by incorporating a large amount of higher-quality data, leading to more stable spectra. In the SSVEP-DAN scheme, significant enhancements in peak amplitudes are observed at the target frequency and its harmonics in both Figure \ref{fig:psd} (a) and Figure \ref{fig:psd} (b). These findings suggest that our proposed SSVEP-DAN method effectively reduces inter-trial variability in SSVEP experiments, enabling the utilization of non-target subject trials and increasing the SNR. Moreover, these phenomena are reflected in the decoding accuracy (Figure \ref{fig:tps}).

\subsection{Limitation and future work}
In the context of SSVEP-DAN, inherent architectural limitations are evident. A key constraint arises from the requirement to match identical stimuli between source and target domains during SSVEP-DAN training, limiting its utility in scenarios involving diverse stimuli. Additionally, the need for a small number of calibration trials for new users restricts its calibration-free usability, affecting user-friendliness.

Regarding future directions, several promising avenues for advancing SSVEP-DAN exist. Firstly, further validation of its effectiveness with state-of-the-art SSVEP classifiers like Compact-CNN \cite{Waytowich2018CCNN}, Conv-CA \cite{li2020conCCA}, DNN \cite{guney2021ssvepDNN}, and SSVEPNet \cite{Pan2022ssvepnet} is essential. Simultaneously, exploring its applicability to other time-locked datasets such as BCI Challenge ERN dataset (BCI-ERN) \cite{margaux2012ERNdataset} can enhance its versatility in time-locked signal decoding. Additionally, researching standardized criteria for source subject selection, considering factors like classification accuracy, SNR, and similarity to target subject, promises to improve SSVEP-DAN's transferability. Lastly, in-depth investigations into the neuroscience underpinnings of model design can provide deeper insights into model interpretation.

\section{Conclusion}
In this study, we propose the first dedicated neural network-based data alignment method, which integrates novel training methods, particularly cross-stimulus training and pre-training techniques. Our experimental results demonstrate that the SSVEP-DAN-based approach enhances subject similarity and improves SSVEP decoding accuracy by effectively utilizing data from non-target subjects. Additionally, the results of ablation studies indicate that through cross-stimulus training and pre-training techniques, further stability in performance enhancement can be achieved. Overall, we expect that our proposed SSVEP-DAN will be applied to an SSVEP-based BCI speller, where the improved SSVEP decoding performance by leveraging the calibration data from a small amount of calibration trials from new subject and the supplementary calibration data from non-target subjects.




\bibliographystyle{IEEEtran}
\bibliography{SSVEP-DAN}

\end{document}